%% file: sample_paper.tex
\documentclass[twoside]{article}

\usepackage[accepted]{aistats2019}
\usepackage{amsmath}
\usepackage{amsfonts}
\usepackage{bm}
\usepackage{amsmath,amsthm,algorithm,algpseudocode}
\usepackage{graphicx}
\usepackage{caption}
\usepackage[outdir=./]{epstopdf}
\usepackage{subfigure} 
\begin{document}

%

%

\twocolumn[

\aistatstitle{Representation Learning on Graphs: A Reinforcement Learning Application}

\aistatsauthor{ Sephora Madjiheurem \And Laura Toni }

\aistatsaddress{ University College London, UK \\ sephora.madjiheurem.17@ucl.ac.uk \And University College London, UK \\ l.toni@ucl.ac.uk} ]
\begin{abstract}
In this work, we study value function approximation in reinforcement learning (RL) problems with high dimensional  state or action spaces via a generalized version of representation policy iteration (RPI).
We consider the limitations of proto-value functions (PVFs) at accurately approximating the value function in low dimensions and we highlight the importance of features learning for an improved low-dimensional value function approximation. 
Then,  we  adopt different representation learning algorithm on graphs to learn the basis functions that best represent the value function.  We empirically show that \textit{node2vec}, an algorithm
for scalable feature learning in networks, and the \textit{Variational Graph Auto-Encoder} constantly outperform the commonly used smooth proto-value functions in low-dimensional feature space.
\end{abstract}

\section{INTRODUCTION}
\input{introduction.tex}


\section{BACKGROUND}
\label{sec:background}
\input{background.tex}

\section{GENERAL REPRESENTATION POLICY ITERATION}
\label{rpi}
\input{methods.tex}

\section{EXPERIMENTS}
\label{sec:result}
\input{experiments.tex}

\section{CONCLUSION}
\label{sec:conclusion}
\input{conclusion.tex}

\bibliography{bibliography}
\end{document}

%% file: introduction.tex
In reinforcement learning (RL), an agent, or decision maker, takes sequential  
actions and  observes the consequent rewards and states, 
which are unknown \emph{a priori}.
These sequent observations improve the agent's knowledge of the environment   with the final goal of learning the optimal policy that maximizes the long term reward. The learning control problem is usually formulated as Markov decision process (MDP), where each state has an associated  value function, which estimates the expected long term reward for some policy (usually the optimal one). Classical MDPs represent the value function by a lookup table, with one entry for each state.  However, this  does not scale with the state (and implicity also action) space dimension, leading to slow learning processes in high-dimensional reinforcement learning problems. Approximated reinforcement learning addresses this problem by learning  a function to properly approximate the true value function.   In the literature, many types of functions have been studied \cite{Kaelbling1996survey, SuttonB98rl}. 

\begin{figure}[t]
      \centering
      \includegraphics[width=1\linewidth]{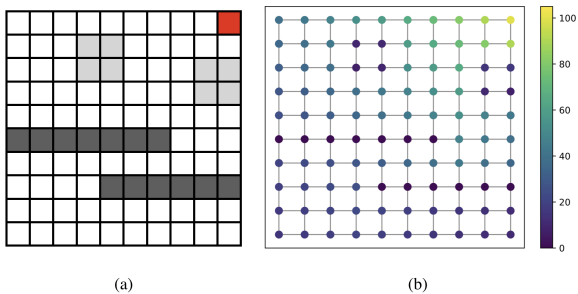}
      \caption{(a) Maze environment. The dark grey squares are strict walls, while the light grey square are difficult access rooms. The red square is the goal room. (b) Optimal value function computed using value iteration \cite{Montague1999}.  }
      \label{fig:maze}
\end{figure}

In this work, we study linear value function approximation, where the value function is  represented as a weighted linear sum of a set of features (called \textit{basis function}). Linear function approximation allows to represent complex value functions by choosing arbitrarily complex basis functions. Under this framework, one of the main challenges is to identify the right set of basis functions. Typical linear approximation architectures such as polynomial basis functions (where each basis function is a polynomial term) and radial basis functions (where each basis function is a Gaussian with fixed mean and variance) have been studied in the case of reinforcement learning \cite{Lagoudakis2003}. These architectures make the assumption that the underlying state space has Euclidiean geometry. However, in realistic scenarios, the MDP's state space is likely to exhibit irregularities.
For instance, let's consider the environment depicted in Figure~\ref{fig:maze}. As it can been seen in Figure~\ref{fig:maze}(b), neighboring states can have values that are far apart (such as states on opposite sides of a wall).
In such cases, these traditional parametric functions may not be able to accurately approximate value functions. 

Consequently, other basis functions have been studied to address this issue. Example of such methods include Fourier basis \cite{Konidaris2011},  diffusion wavelets, \cite{Mahadevan2006diffusionwavelets}, Krylov basis  \cite{PetrikTPZ2010kyrlo} and Bellman Error Basis Function \cite{Parr2007featureGen, Parr2008featureselecRL}. 
In particular, work by \cite{Mahadevan2007} introduces the \textit{representation policy iteration} (RPI), a  spectral graph framework for solving Markov decision processes by jointly learning representations and optimal policies. In their work, the authors first note that MDPs can be intuitively  represented using graphs, with states being the nodes and the transition probability being the similarity matrix. Then, under the assumption that the value function is usually modelled as a diffusion process over the state-graph (and therefore it is a smooth function), they approximate the value function (smooth signal on the graph) as a linear combination of the first Laplacian eigenmaps on the state-graph. These features, known as \textit{proto-value functions} (PVFs), preserve the smoothness of the value function. 
In this paper, we argue that constructing the graph that perfectly models the MDP such that the value function is indeed smooth on the graph is not trivial. Therefore, there is a need to automatically learn the basis functions that capture the geometry of the underlying state space from limited data to further improve the performance.
Hence, given the success of recent node embedding models \cite{Grover2016, KipfW16vgae, Ribeiro2017struc2vec, Donnat2018graphwave}, we propose to   investigate representation learning on graphs algorithms to learn basis functions in the linear value function approximation. 

The idea behind recent successful representation learning approaches is to learn a mapping that embeds the nodes of a graph as low-dimensional vectors. They aim to optimize the representations so that geometric relationships in the embedding space preserve the structure of the original graph. \cite{Hamilton2017RLongraph} surveys recent representation learning on graph methods. Therefore, in this work, we generalize the RPI algorithm \cite{Mahadevan2007} to allow different basis functions, and analyze the performance of several representation learning methods for value function approximation. 

The rest of this paper is structured as follows: in Section~\ref{sec:background} we introduce background material, providing details on Markov decision processes and value function approximation and describe the representation learning algorithm used in this work. The General Representation Policy Iteration algorithm is described in Section~\ref{rpi}. In Section~\ref{sec:result}, we discuss experimental results and proceed to summarize the main findings and give direction for future work. We finally conclude in Section~\ref{sec:conclusion}.

%% file: background.tex
\subsection{Markov decision process (MDP)}
Markov decision processes are discrete time stochastic control processes that provide a widely-used mathematical framework for modeling decision making strategies under uncertainty.   
Specifically, at each time step, the process is in some state $s$, and the agent can choose any action $a$ that is available in state $s$. As a consequence of  the action taken,  the agent finds itself in a new state $s'$ and  observes an instantaneous reward $r$.
We define a discrete MDP by the tuple $M = (S, A, P, R)$, where $S$ is a finite set of discrete states, $A$ a finite set of actions, $P$ describes the transition model $-$with $P(s,a,s')$ giving the probability of moving from state $s$ to $s'$ given action $a$ $-$ and $R$ describes the reward function $-$ with $R(s,a)$ giving the immediate reward from taking action $a$ in sate $s$. 
Given a policy $\pi : S \mapsto A$, a value function $V^{\pi}$ is a mapping $S \mapsto \mathbb{R}$ that describes the expected long-term discounted sum of rewards observed by the agent in any given state $s$ when following policy $\pi$. Solving a MDP requires to find a policy that defines the optimal value function $V^*$, which satisfies the following constraints:
\begin{equation*}
V^*(s) = \max_a \Big( R(s,a) + \gamma \sum_{s'\in S} P(s,a,s') V^*(s') \Big).    
\end{equation*}
 This recursive equation is known as the \textit{standard form} of Bellman's equation.  The optimal policy is a  unique solution to the Bellman's equation and can be found by dynamic programming, iteratively evaluating the value functions for all states. 
\subsection{Value Function Approximation}
In large state spaces, computing exact value functions can be computationally intractable. A possible solution  is to estimate the value function with function approximation (value function approximation method) \cite{Bertsekas1996ndp}. Commonly, the value function is approximated as a weighted sum of a set of features (called \textit{basis function}) \cite{Montague1999, Mahadevan2007b,Konidaris2011,Lagoudakis2003}: 
\begin{equation*}
\phi_1, \phi_2, \ldots, \phi_d  :~ \tilde{V}(s| \theta)~=~\sum_{i = 1}^d \theta_i \phi_i(s)~\approx~V(s),
\end{equation*}
where $d$ is the dimension of the features space.

The basis functions $\phi_i$ can be hand-crafted \cite{SuttonB98rl} or automatically constructed \cite{Mahadevan2007}, and the model parameters $\bm \theta = [\theta_1,\theta_2, \ldots, \theta_d]$ are typically learned via standard parameter estimation methods such as least-square policy iteration (LSPI) \cite{Lagoudakis2003}. However, how to properly design the set of basis for a data-efficient function approximation framework is still an open question. The main question is how  to find the set of basis $\bm \phi$ that is \textit{low-dimensional} (to ensure a data-efficient learning) and yet a \textit{meaningful representation} of the MPD (to reduce the suboptimality due to the  value function approximation).

The representation policy iteration algorithm (RPI) was introduced in \cite{Mahadevan2007} to address this problem. It is a three steps algorithm consisting of (1) a sample collection phase, (2) a representation learning phase and (3) a parameter estimation phase. RPI is described in further details in Section~\ref{rpi}. 
In this work, we propose to generalize RPI to allow different representation learning methods. In particular, we first observe that state space topologies of MDPs can be intuitively modeled as (un)directed weighted graphs, with the nodes being the states and the transition probability matrix being the similarity matrix. When the transition probabilities are unknown, we can construct a graph from collected samples by connecting temporally consecutive states with a unit cost edge. Therefore, similarly to \cite{Mahadevan2007b}, we propose to construct the graph from collected samples of an agent acting in the environment given by the MDP. We then learn representations on the graph induced by the MDP using node embedding methods. Finally, we use the learned representations to linearly approximate the value function. In the next section, we describe the node embedding models that we exploit within this framework.

\subsection{Representation Learning on Graph}
\label{reprLearning}
We propose to use the following learned node embedding models as basis functions for the value function approximation in order to automatically learn to encode the graph structure - hence the MDP - into low-dimensional embeddings. 
\paragraph{Node2Vec}
\textit{Node2vec} \cite{Grover2016} is an algorithmic framework for learning continuous feature representations for nodes in networks. It is inspired by the powerful language model \textit{Skip-gram} \cite{mikolov2013word2vec} which is based on the hypothesis that words that appear in the same context share semantic meaning. In networks, the same hypothesis can be made for nodes, where the context of a node is derived by considering the nodes that appear in the same random walk on the graph. Therefore,
node2vec learns the node embeddings based on random walk statistics. The key is to optimize the node embeddings so that nodes have similar embeddings if they tend to co-occur on short (biased) random walks over the graph. Moreover, it allows for a flexible definition of random walks by introducing parameters that allow to interpolate between walks that are more breadth-first search or depth-first search.

Specifically, for a graph $G = \mathcal{(V,E}, W)$ (where $\mathcal{V}$ is a set of nodes, $\mathcal{E}$ a set of edges and $W$ the weight matrix) and a set $\mathcal{W}$ of $T$ biased random walks collected under a specific sampling strategy on the graph $G$,
node2vec seeks to maximize the log-probability of observing the network neighborhood of each node $u \in \mathcal{V}$ conditioned on its features representations, given by $f$ (a matrix of size $|V| \times d$ parameters, where $d$ is the dimension of the feature space):
\begin{equation*}
\max_f \sum_{w \in \mathcal{W}} \sum_{t = 1}^T \log Pr(N_{w}(u_i) | f(u_i)),     
\end{equation*}
where $N_{w}(u_i)$ describes the neighborhood of the $i$th node in the walk $w$.
\paragraph{Struc2Vec}
By introducing a bias in the sampling strategy, node2vec allows to learn representations that do not only focus on optimizing node embeddings so that nearby nodes in the graph have similar embeddings, but also consider representations that capture the structural roles of the nodes, independently of their global location on the graph.
The recent node embedding approach, \textit{struc2vec}, proposed by \cite{Ribeiro2017struc2vec} addresses the problem of specifically embedding nodes such that their structural roles are preserved. 
The model generates a series of weighted auxiliary graphs $G_{k}$ (with $k = {1,2,...}$) from the original graph $G$, where the auxiliary graph $G_{k}$ captures structural similarities between nodes $k$-hop neighborhoods. Formally, let $R_k(u_i)$ denote the ordered sequence of degrees of the nodes that are exactly k-hops away from $u_i$, the edge-weights, $w_k(u_i,v_j)$, in the auxiliary graph $G_{k}$ are recursively represented by the structural distance between nodes $u_i$ and $v_j$ defined as
\begin{equation*}
 w_k(u_i,v_j) = w_{k-1}(u_i,v_j)+d(R_k(u_i),R_k(u_j)),   
\end{equation*}

where $w_0(u_i,v_j) = 0$ and $d(R_k(u_i),R_k(u_j))$ is the distance between the ordered degree sequences $R_k(u_i)$ and $R_k(u_j)$ computed via dynamic time warping \cite{Ribeiro2017struc2vec}. 

Once the weighted auxillary graphs $G_{k}$ are computed, struc2vec runs biased random walks over them and proceeds as node2vec, optimising the log-probability of observing a network neighborhood based on these random walks.

\paragraph{GraphWave}
The \textit{GraphWave} algorithm as proposed by \cite{Donnat2018graphwave} takes a different approach to learning structural node embeddings. It learns node representations based on the diffusion of a spectral graph wavelet centered at each node.
For a graph $G$, $\mathcal{L} = D - A$ denotes the graph Laplacian, where $A$ is the adjacency matrix and $D$ is a diagonal matrix, 
whose entries are row sums of the adjacency matrix. Let $U$ denote the eigenvector decomposition of the graph Laplacian $\mathcal{L} = U \Lambda U^T$ and $\Lambda = \text{diag}(\lambda_1, \lambda_2, \ldots, \lambda_{|V|})$ denote the eigenvalues of $\mathcal{L}$. Given a heat kernel $g_s(\lambda) = e^{-s\lambda}$ for a given scale $s$, GraphWave uses $U$ and $g_s$ to compute a vector $\bm \psi_{u}$ representing  diffusion patterns for node $u$ as follows:
\begin{equation*}
\bm \psi_{u} = U \text{diag}(g_s(\lambda_1), g_s(\lambda_2), \ldots, g_s(\lambda)_{|V|}) U^T \delta_{u} 
\end{equation*}

where $\delta_{u} $ is the one-hot indicator vector for node $u$.
Then, the characteristic function for each node's coefficients $\bm \psi_{u}$ is computed as
\begin{equation*}
    \phi_{u}(t) = \frac{1}{|V|} \sum_{m = 1}^{|V|} e^{it\Psi_{mu}}
\end{equation*}
Finally, to obtain the structural node embedding $f(u)$ for node $u$, the paramatric function $\phi_{u}(t)$ is sampled at $d$ evenly spaced points $t_1, \ldots, t_d$:
\begin{equation*}
f(u) = \big[\text{Re}(\phi_{u}(t_i),\text{Im}(\phi_{u}(t_i))\big]{t_1,\ldots,t_d}.
\end{equation*}
\paragraph{Variational Graph  Auto-Encoder}
As opposed to directly encoding each node, auto-encoders aim at directly incorporating the graph structure into the encoder algorithm. The key idea is to compress information about a node's local neighborhood. The \textit{Variational Graph  Auto-Encoder} proposed by \cite{KipfW16vgae} is a latent variable model for graph-structure data capable of learning interpretable latent representations for undirected graphs.
The Graph Auto-Encoder uses a Graph Convolutional Neural Network (GCN) \cite{KipfW16gcn} to encode graphs and another GCN to reconstruct the graph. The Variational Graph Auto-Encoder makes use of latent variables. 

%% file: methods.tex
Within the context of approximated value function, the representation policy iteration algorithm (RPI) was introduced in \cite{Mahadevan2007} to  learn the approximating function. RPI is a three step algorithm consisting of (1) a sample collection phase,  to build a training dataset with  quadruples  $ \{(s_i, a_i,s_{i+1}, r_i)\}$;  (2) a representation learning phase that defines a set of basis functions; and (3) a parameter estimation phase, in which the coefficients of the linear approximation are learned.  A generalized version of the RPI algorithm \cite{Mahadevan2007b} is described in Algorithm~\ref{alg:fig}. 

\begin{algorithm}
\caption{General Representation Policy Iteration}
\label{alg:fig}
\begin{algorithmic}
\State \textbf{Input:} \\$\pi_0$: sampling strategy, \\$N$: number of random walks to sample, \\$T$: length of each walk, \\$d$: dimension of the basis functions, \\$model$:  representation learning model, \\$\epsilon$: convergence condition for LSPI.

\State \textbf{Output:} $\epsilon$-optimal policy $\pi$
\State \textbf{1. Sample Collection Phase}
\State Collect a data set $\mathcal{D}$ of $T$ successive samples $ \{(s_i, a_i,s_{i+1}, r_i),(s_{i+1}, a_{i+1},s_{i+2}, r_{i+1}), \ldots\}$ by following sampling strategy $\pi_0$ for maximum $T$ steps (terminating earlier if it results in an absorbing goal state).
\State \textbf{2. Representation Learning Phase}
\State Build basis function matrix $\bm \phi = model(\mathcal{D}, d)$.
\State \textbf{3. Control Learning Phase}
\State Using a parameter estimation algorithm such as LSPI or Q-learning, find an $\epsilon$-optimal policy $\pi$ that maximizes the action value function $Q^{\pi} = \bm \phi \theta^{\pi}$ within the linear span of the basis $\bm \phi$.
\end{algorithmic}
\addtocounter{algorithm}{-1}
\end{algorithm}

In the original RPI, the representation learning phase is predefined. Namely, an undirected weighted graph $G$ is built from  the available data set  $\mathcal{D}$. Then a diffusion operator $O$, such as the normalized Laplacian is computed on graph $G$ and the $d$-dimensional basis functions   $\pmb{\phi}=[\phi_1, \ldots, \phi_d]$ are constructed from spectral analysis of the diffusion operator. Specifically, the $\phi_i$'s are the smoothest eigenvectors of the graph Laplacian and are known as \textit{proto-value functions} (PVFs). 
The key is that given a state-graph that perfectly represents the MDP, the value function is modelled as a diffusion process over the graph (and therefore it is a smooth function). Hence, given the spectral properties of the Laplacian operator, the proto-value functions are a good choice of basis functions for preserving the smoothness of the value function. 

However, it is not guaranteed  that we can construct a graph from a limited number samples such that its derived proto-value functions reflect the underlying state space. If fact, we can show that the value function is not as smooth on the estimated graph (constructed from samples) as it is on the ideal graph where the edges are weighted by the transition probability. 
We consider the environment depicted in Figure~\ref{fig:maze}. To construct the estimated graph $\hat{G}$, we first collect samples by running 100 independent episodes starting at a random initial state and taking successive random actions until either a maximum of 100 steps have been made or the goal state has been reached. We then connect temporally consecutive states with a unit cost edge.
The ideal graph $G$ is simply the graph with edges representing actual transition probabilities (i.e. edges between accessible states have weight 1, edges between an accessible state and a wall state have weight 0, and edges between an accessible or difficult access state and a difficult access state have weight 0.2).
We use the following function to measure the global smoothness of the value function on a graph: 

\[
\sum_{i,j \in \mathcal{E}} w_{ij} (v_i - v_j)^2 = \bm{v^T} \mathcal{L} \bm{v}.
\]

Where $\mathcal{L}$ is the graph Laplacian. In other words, if values $v_i$ and $v_j $ from
a smooth function reside on two well connected nodes (i.e. $w_{ij}$ is large), they are expected to have a small distance $(v_i - v_j)^2$, 	hence $\bm{v^T} \mathcal{L} \bm{v}$ is small overall.

As seen in Table~\ref{tab:smoothness}, this analysis shows a reduction of the value function smoothness when going from the ideal weighted graph to the estimated and unweighted graph (usually considered in realistic settings, when the transition probability is not known a priori). 

\begin{table}[ht]
\centering
\begin{tabular}{|l|c|}
\hline
                & $\bm{v^T} \mathcal{L} \bm{v}$     \\ \hline
Estimated graph & 14831.72 \\ \hline
Weighted graph  & 5705.65  \\ \hline
\end{tabular}
\caption{Analysis of the smoothness of the value function on different graphs.}
\label{tab:smoothness}
\end{table}

\begin{figure}[ht]
      \centering
      \includegraphics[width=1\linewidth]{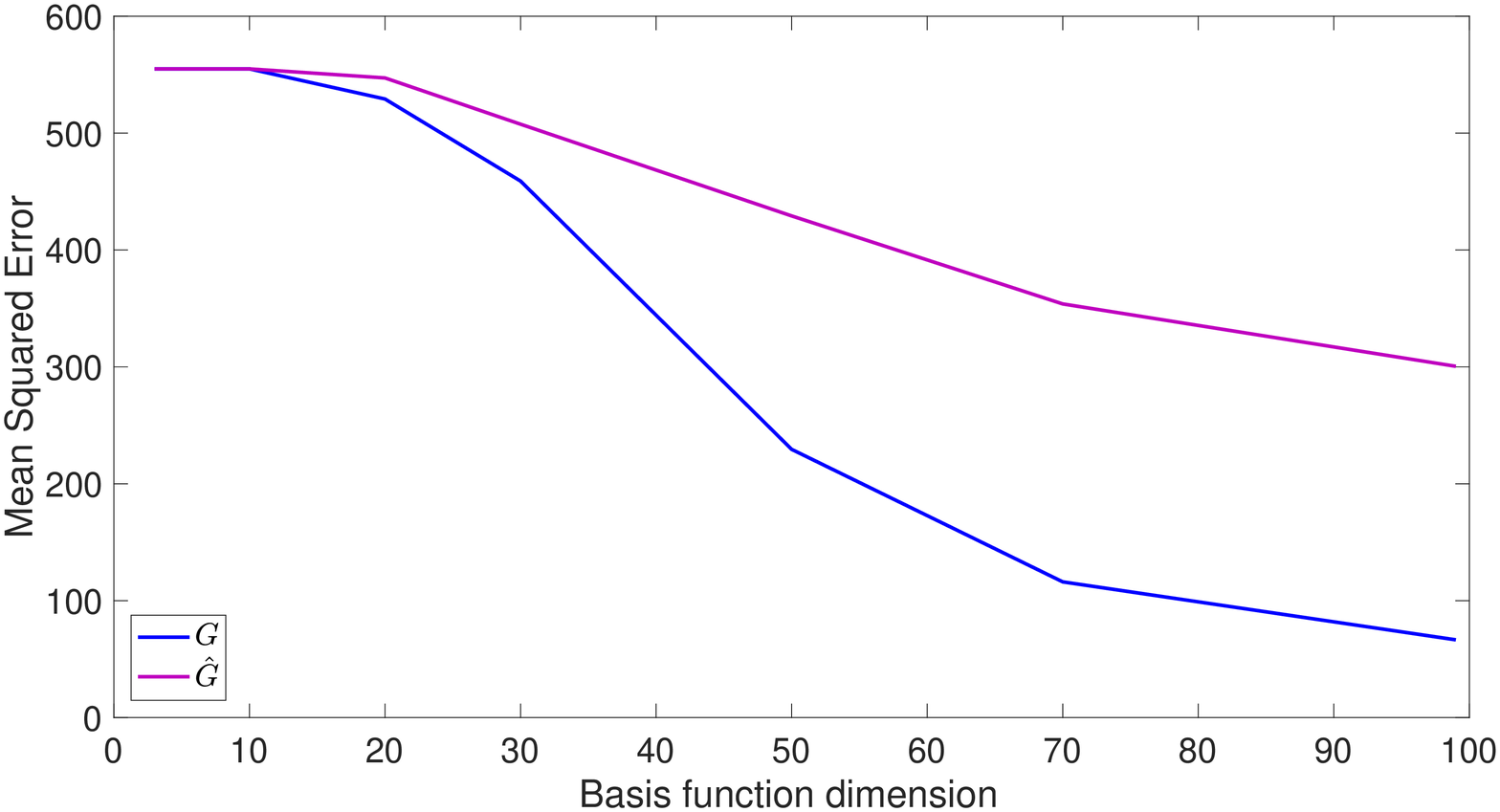}
      \caption{ MSE between the approximated value function and the true value function when using proto-value functions generated from two different graphs as basis functions. On the $x$ axis we make the dimension of the basis function (the number of proto-value functions) vary. Best seen in color.}
      \label{fig:smoothness}
\end{figure}

As results, it is expected that the smoothest proto-value functions of the estimated graph $\hat{G}$ on which the value function is less smooth, do not allow to reconstruct the true value function as well as the smoothest proto-value functions of the ideal graph. This phenomenon is verified in Figure~\ref{fig:smoothness}, where we show in both cases the mean squared error (MSE) of the approximate value function computed in a least-square way using the true value function computed via value iteration \cite{Montague1999} for the environment shown in Figure~\ref{fig:maze}.

To overcome this limitation, we propose to use the node embedding methods described in Section~\ref{reprLearning} to automatically learn the basis functions from the geometry of the underlying state space to further improve the performance.
In the following, we describe how to apply these features learning methodologies within reinforcement learning strategies.

%% file: experiments.tex
\subsection{Set up}
\begin{figure}[!t]
\center
\subfigure[]{\includegraphics[width=1.5in]{{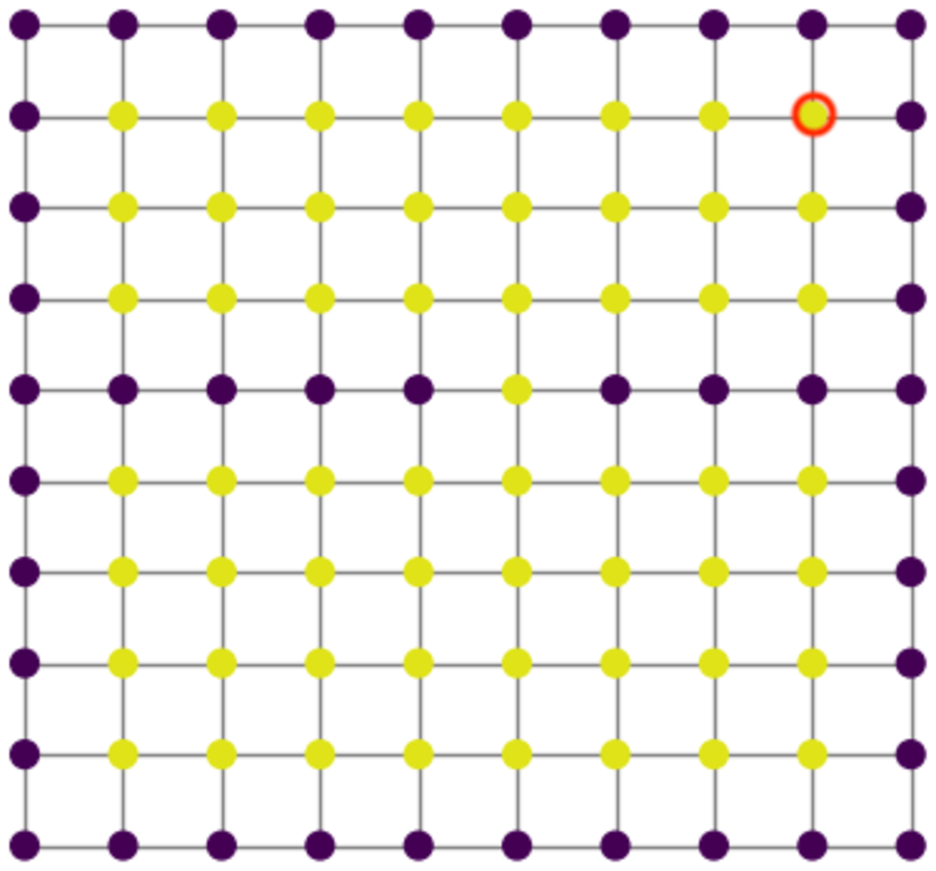}} \label{fig:tworooms}}
\subfigure[]{\includegraphics[width=1.5in]{{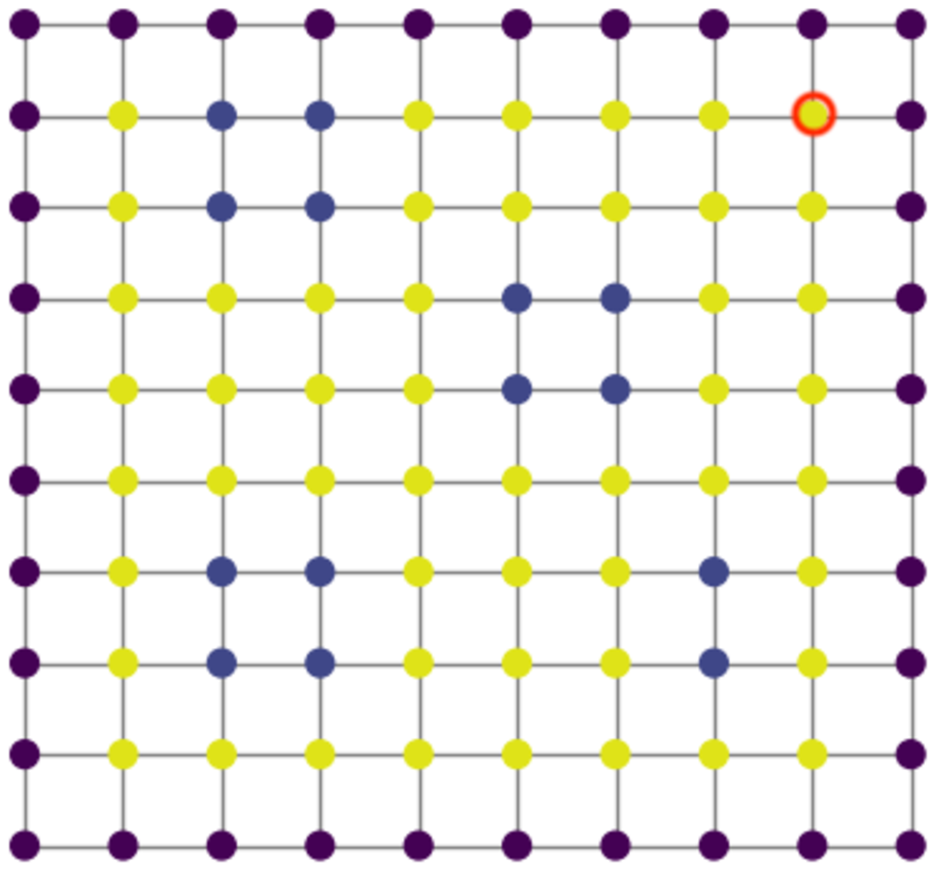}} \label{fig:obstacle}}
\hfill
\caption{ Two different maze environments. The purple nodes represent strict walls, while the blue nodes are difficult access rooms. All other nodes represent accessible rooms The node shown in red is the goal room. Best seen in color.}
\end{figure}

We consider the two-room environment used in \cite{Mahadevan2007}, shown in Figure~\ref{fig:tworooms}. It consists of 100 states in total, divided into 57 accessible states and 43 inaccessible states representing walls. There is one goal state, marked in red and the agent is rewarded by $+100$ for reaching the goal state.

We also consider the obstacles-room environment depicted in Figure~\ref{fig:obstacle}. In this environment, there are $100$ states in total, some of which are inaccessible since they represent exterior walls and $14$ of which are accessible from neighbouring states with a fixed probability of $0.2$ (they represent a moving obstacle or difficult access space). All the other states are reachable with probability $0.9$.  The agent is rewarded by $+100$ for reaching the state located at the upper-right corner.  

We construct the corresponding graphs where each location is a node, and the transitions (4 possible actions: left, right, up and down) are represented by the edges. 

We run and evaluate the General Representation Policy Iteration (GRPI) algorithm using embedding models from Section~\ref{reprLearning} to compute the basis functions in the second phase of the algorithm.

\begin{figure}[!t]
\center
\subfigure[]{\includegraphics[width=3.5in]{{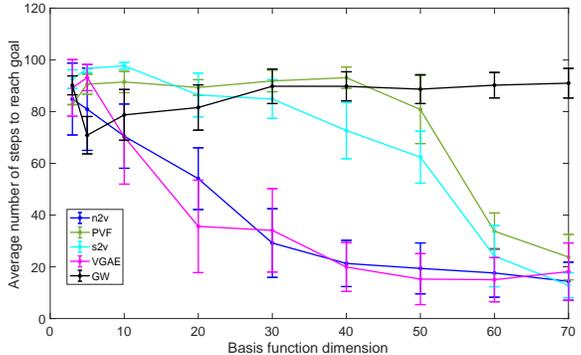}} \label{fig:tworooms_results}}
\subfigure[]{\includegraphics[width=3.5in]{{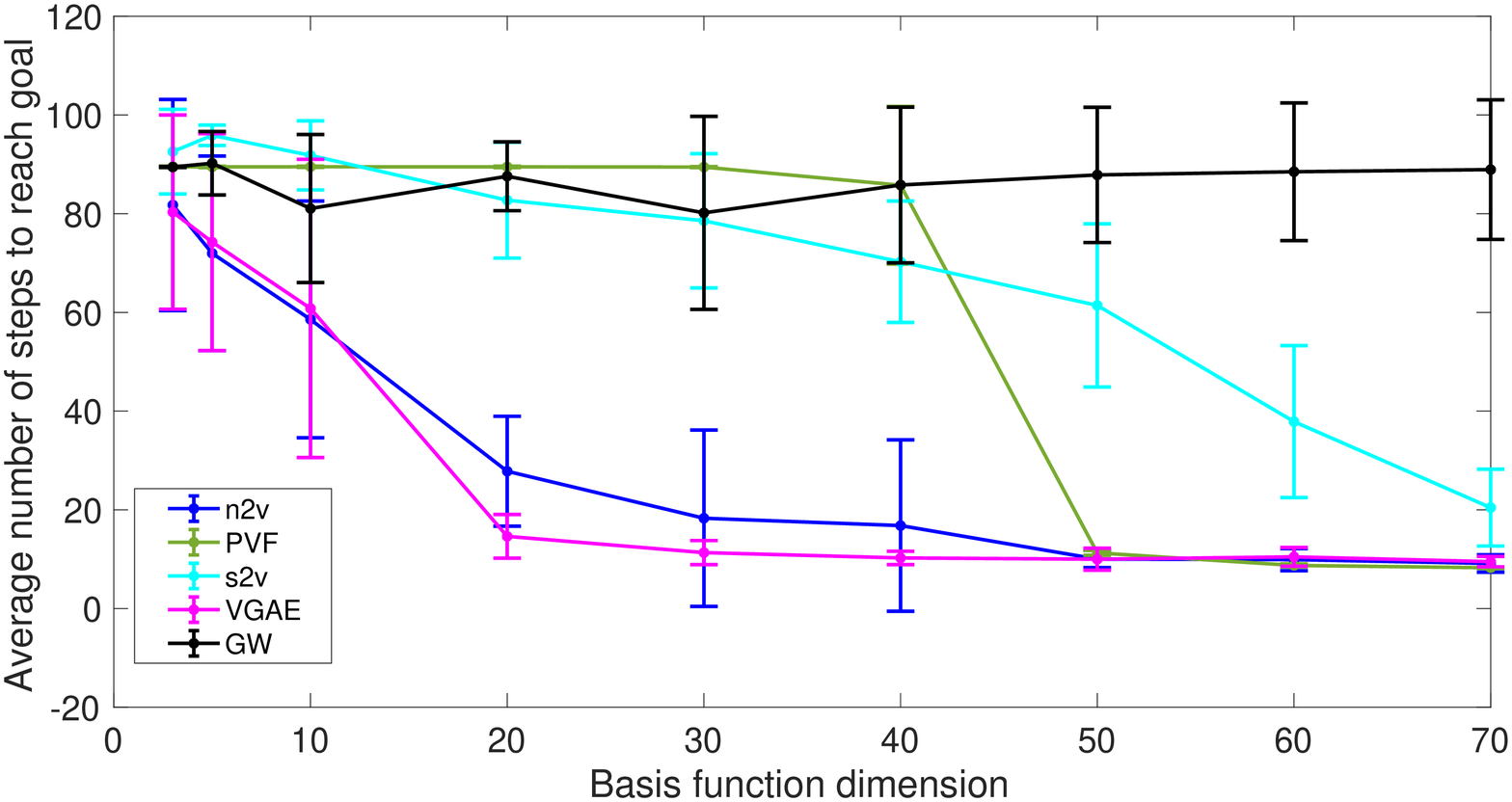}} \label{fig:obstacle_results}}
\hfill
\caption{ Average number of steps required to reach the goal steps using the various basis functions. On the $x$ axis we make the dimension of the basis functions vary. Best seen in color. \ref{fig:tworooms_results} corresponds to the two-room environment, while \ref{fig:obstacle_results} corresponds to the obstacle-room environment.}
\end{figure}

\begin{enumerate}
    \item We first collect a set $\mathcal{D}$ of 100 sampled random walks, each of length 100 (or terminating early when the goal state was reached). The sampling dynamic is as follows:  starting from a random accessible sate, the agent takes one of the four possible actions (move up, down, left or right). If a movement is possible, it succeeds with probability 0.9. Otherwise, the agent remains in the same state. If the agent reaches the gold state, it receives a reward of 100, and is randomly reset to an accessible interior state. We use off-policy sampling ($\pi_0 =$ random policy) to collect the samples, except in the case of node2vec, where the samples are generated under a biased random walk. We use grid search to find the optimal hyperparameters $p = 1$ and $q = 4$ that guide the walk according to \cite{Grover2016}.
    \item We then use sample transitions in $\mathcal{D}$ to build an undirected graph where the weight matrix $W$ is the adjacency matrix and run $model(\mathcal{D}, d)$ with $model \in $ \{node2vec (n2v), struc2vec (s2v), variational graph auto-encoder (VGAE), GraphWave (GW)\} for diffenrent choices of $d$. In the case of node2vec, we reuse the samples set to derive the node neighbourhoods used in the objective function. 
    \item We learned the parameters of the linear value approximation using the parameter estimation method LSPI \cite{Lagoudakis2003} with the set of samples $\mathcal{D}$.
    \item We used the policies learned by GRPI for each model to run simulations starting from each accessible states. We compare the performance of each models in terms of the average number of steps required to reach the goal. We also compare to the traditional PVF basis functions. The results for the two environments, averaged over 20 independent runs, are shown in Figures~\ref{fig:tworooms_results} and ~\ref{fig:obstacle_results}. Each run consists of episodes of a maximum of 100 steps, where each episode is terminated earlier if the agent reached the goal state.
\end{enumerate}
\subsection{Discussion}
\begin{figure}[ht]
      \centering
      \includegraphics[width=2.5in]{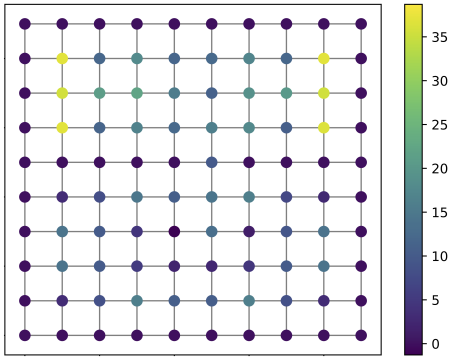}
      \caption{Approximate value function via GRPI using GraphWave basis function of dimension 70 on the two-room environment.}
      \label{fig:gwapprox}
\end{figure}

Figures~\ref{fig:tworooms_results} and ~\ref{fig:obstacle_results} show the average number of steps to reach the goal as a function of the dimension of the basis function.  We first observe that the policy learned via the GraphWave basis function lead to very poor performances regardless of the dimension size. We investigate this phenomenon by looking at the approximate value function learned under these basis. 
The approximate state values are depicted in Figrue~\ref{fig:gwapprox}. Because GraphWave aims at learning embeddings that are exclusively structural, we hypothesise that they fail at capturing global network properties. In fact, the embeddings learned by GraphWave for the corner states in the two-room environment are equals, making it obviously impossible to learn different state values with linear approximation. This suggests that although the GraphWave is a powerful model for capturing structural information in networks, it is not a good choice of basis function for approximating value function on a graph.

On the other hand, we notice that although struc2vec was also designed to capture structural similarities between nodes, it also preserves the local properties of the graph by considering neighborhoods of different sizes \cite{Ribeiro2017struc2vec}. Hence, struc2vec is able to accurately approximate the value function even in graphs that have symmetrical structure such as the two-room environment. 

Finally, the result show that VGAE and node2vec are good choices of basis functions for approximating the value function in low dimension. Indeed, they lead to good performances in terms of number of steps to reach the goal states with basis functions of dimension as low as 20 for VGAE and 30 for node2vec. On the contrary, we observe that the PVFs require dimension of at least 70 to reach comparable performances on the two-room domain and dimension of 50 on the obstacle-room domain. 

\begin{figure}[t]
      \centering
      \includegraphics[width=3.5in]{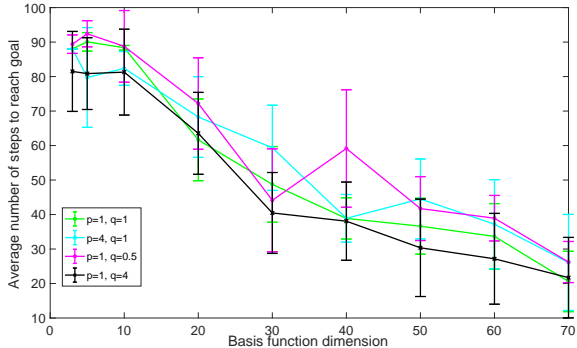}
      \caption{Average number of steps required to reach the goal steps using node2vec with varying parameters $p$ and $q$. On the $x$ axis we make the dimension of the basis functions vary.}
      \label{fig:n2vsampling}
\end{figure}

We observed that the sampling strategy used in node2vec has a significant impact on the performance of the learned policy. Using grid search, we find that the optimal value of the parameters $p$ and $q$ that guide the random are $1$ and $4$ respectively. We show the performances of node2vec with selected values of $p$ and $q$ in Figure~\ref{fig:n2vsampling}.
When $p < q$ and $q > 1$, the strategy is biased to encourage walks to backtrack a step and to visit nodes that are close to the current node in the walk. Therefore, it leads to walks that approximate a breadth-first search behavior, gathering a local view of the underlying graph with respect to the starting node. On the other hand, when $p > q$ and $q < 1$, the walk approximate a depth-first search behavior and lead to more outward exploration.  \cite{Grover2016} show that the former type of sampling strategy allows to reflect structural equivalences of nodes whereas the second type allows to capture homophily within the network. Figure \ref{fig:n2vsampling} suggests that for approximating value functions, structural equivalence plays a more important role than homophily.   

\subsection{Additional Results}

In order to investigate whether we can expect similar a behaviour in larger environments, we consider a 100 by 50 three-room environment (similar to the two-room environment but with two interior walls, with the upper wall having the opening more on the right and the lower wall having the opening more on the left). We construct the graph from 500 collected samples of length at most 100 and derive the PVFs and the node2vec embeddings. 
For each of these  basis function, we solve the linear approximation problem in the least-square sense by minimizing the following loss function with respect to the parameter $\bm \theta$ using the optimal value function computed via value iteration \cite{Montague1999}:

\begin{equation*}
L(\theta) = \frac{1}{|S|}\sum_{s \in S} \Big( V(s) - \sum_{i = 1}^d \theta_i \phi_i(s) \Big)^2. 
\end{equation*}

\begin{figure}[t]
      \centering
      \includegraphics[width=3.5in]{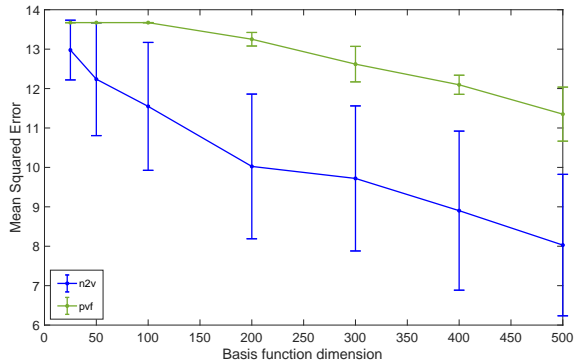}
      \caption{Mean squared error of value function approximation. On the $x$ axis we make the dimension of the basis functions vary.}
      \label{fig:threerooms}
\end{figure}

Figure`\ref{fig:threerooms} shows the gain of adopting node2vec feature learning in reinforcement learning in high dimensional state space.

\subsection{Main Findings and Future Work}
We summarize below the main findings of our work. 
\begin{itemize}
	\item The smoothness assumption of the value function on an estimated unweighted graph does not necessarily hold. 
    \item Using basis functions that automatically learn to embed the geometry of the graph induced by the MPD can lead to improved performance over the proto-value functions.
    \item Such embedding models need to capture the structural equivalence of the nodes while preserving the local properties of the graph.
    \item Under sampling strategies that satisfy the requirements of the previous point, Node2vec \cite{Grover2016} outperforms the commonly used proto-value functions.
    \item The Variational Graph Auto-Encoder, which is more complex than node2vec and requires more training, leads to minor performance improvement compared to node2vec.
\end{itemize}

These findings encourage the further study of representation learning on graphs for achieving efficient and accurate policy learning for reinforcement learning. In particular, the question of scalability in large or continuous state space arises. Future work includes analyzing to what extend one can efficiently learn good embeddings with limited samples in very large state spaces. Another interesting open question in this direction, is to investigate whether good representations can be inferred for states that have never been visited.

Naturally, future work should also aim at further improving the quality of the embeddings for solving reinforcement learning problems. A possibility would be to make use of the reward observed during the sample collection phase to build features that are not only based on state transitions, but capture reward information as well.

%% file: conclusion.tex
In this work, we have studied the representation policy iteration algorithm with a modified representation learning phase that allows to use any model for computing the basis functions in the linear value approximation. We investigate several models for learning high quality node embeddings that preserve the geometry of the graph induced by the Markov decision process. We compare the performance of several representation learning model in the context of value function approximation. Finally, we observe that models that are designed to capture the global structural geometry of the graph while preserving local properties do well at approximating the value function in low feature space dimensions, significantly outperforming the commonly considered proto-value functions for this task.